\documentclass[twoside,11pt]{article}
\usepackage[preprint]{jmlr2e}
\usepackage[T1]{fontenc}
\usepackage{lmodern}
\usepackage{array}
\usepackage{booktabs}
\usepackage{listings}
\usepackage{tikz}
\usetikzlibrary{arrows.meta,positioning,fit,calc,shadows}
\usepackage{xcolor}
\newcolumntype{L}[1]{>{\raggedright\arraybackslash}p{#1}}
\jmlrheading{23}{2026}{1--??}{1/21; Revised 5/22}{9/22}{21-0000}{J. Zhang, Larionov, Wang, Yin, and W. Zhang}

\ShortHeadings{FairLMs: A Turnkey Library for Fairness in Language Models}{J. Zhang, Larionov, Wang, Yin, and W. Zhang}
\firstpageno{1}

\begin{document}

\title{FairLMs: A Turnkey Library for Fairness in Language Models}

\author{\name Jiale Zhang \email jz106@iu.edu \\
       \addr Department of Biostatistics and Health Data Science\\
       Indiana University\\
       Indianapolis, IN, United States
       \AND
       \name Michael Larionov
       \email mlariono@andrew.cmu.edu \\ \addr Department of Statistics and Data Science\\Carnegie Mellon University\\Pittsburgh, PA, United States
       \AND
       \name Zichong Wang \email zwang114@fiu.edu \\
       \name Zhipeng Yin \email zyin007@fiu.edu \\
       \name Wenbin Zhang \textsuperscript{*}\email wenbin.zhang@fiu.edu \\
       \addr Knight Foundation School of Computing and Information Sciences\\
       Florida International University\\
       Miami, FL, United States}


\makeatletter
\begingroup
\renewcommand{\thefootnote}{*}
\renewcommand{\@makefntext}[1]{%
  \noindent\@makefnmark\hspace{0.5em}#1%
}
\footnotetext{Corresponding author}
\endgroup
\makeatother

\maketitle

\vspace{-0.6cm}
\begin{abstract}
Fairness research on language models involves measuring bias, applying mitigation methods, and examining the evidence on which an evaluation rests. Existing tools offer complementary functionality through different interfaces, so combining them requires reconciling model interfaces, evidence formats, access constraints, and result types before applicability can be checked or methods compared. We introduce \textbf{FairLMs}, a Python library that connects these activities through explicit declarations of model capabilities and input requirements. It provides 33 intrinsic and extrinsic metrics, 14 mitigation components spanning four intervention categories, 14 dataset and scoring-instrument diagnostics, adapters for the three Transformer architectures and supported hosted completion APIs, and benchmark loaders. Declarations are checked before execution and results carry the configuration under which they were obtained, so that compatible components can be combined, methods compared under a common protocol, and workflows extended to new models and datasets. The source code is available at \url{https://github.com/FairLMs/FairLMs}.
\end{abstract}


\vspace{-0.35cm}
\section{Introduction}
\label{sec:intro}
\vspace{-0.25cm}

Language models (LMs) are increasingly deployed in high-stakes domains such as healthcare, finance and employment~\citep{lee2026clash,luo2025leveraging,niszczota2023gpt,gan2024application}, where biased behavior can harm already underserved populations. A fairness study of such a model requires three decisions that are ordinarily made with separate toolchains: the choice of metric, which depends on the model architecture and on whether bias is measured intrinsically, in internal representations, or extrinsically, in downstream behavior~\citep{gallegos2024biasfairness}; the choice of mitigation method, which depends on its intervention category and required access level~\citep{zhang2026mitigating}; and an assessment of whether the evaluation dataset and the auxiliary scoring instrument support the intended interpretation of the result~\citep{blodgett2021stereotyping}. These decisions are interdependent: a mitigated model must be re-measured under the same metrics, and the interpretation of any measurement depends on the dataset and the scoring instrument that produced it.

However, existing software covers this workflow only in parts. General-purpose fairness toolkits~\citep{bellamy2019ai,weerts2023fairlearn}, evaluation harnesses~\citep{eval-harness,liang2022holistic,bouchard2025langfair}, paper-specific debiasing collections~\citep{meade2022empirical}, LM bias toolkits~\citep{viswanath2023fairpy} and embedding-level libraries~\citep{badilla2025wefe} each address part of it. Combining them therefore requires locating a separate implementation for each metric, adapting it to the tokenizer and output-layer conventions of each model, repeating that adaptation for every additional model, and implementing a dedicated procedure for comparing a mitigated system with its baseline.  To address this, we introduce \textbf{FairLMs}, a library that integrates the three activities under a common interface: 33 metrics, 14 mitigators and 14 dataset and scoring-instrument diagnostics. Each component declares the architectures, model capabilities, and evidence containers it works with, and these declarations are checked before execution. Unsupported combinations are thus caught before any computation, a model adapted once serves every component it satisfies, and each result records the configuration that produced it, so reported values can be interpreted and reproduced.

\vspace{-0.45cm}
\section{Library Design}
\label{sec:library}
\vspace{-0.15cm}

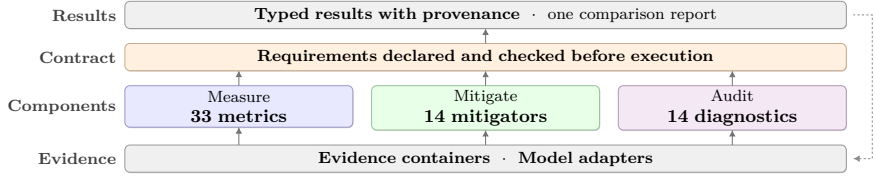
\begin{figure}[t]
\vspace{-0.35cm}
\centering
\resizebox{0.75\linewidth}{!}{%
\begin{tikzpicture}[
  font=\scriptsize,
  every node/.style={inner sep=0pt, outer sep=0pt},
  band/.style 2 args={draw=#1, fill=#2, rounded corners=2.5pt, line width=0.35pt,
    text width=12.5cm, align=center, minimum height=4.6mm, inner ysep=1.1mm,
    drop shadow={shadow xshift=0.35pt, shadow yshift=-0.5pt, opacity=0.18, fill=black}},
  cell/.style 2 args={draw=#1, fill=#2, rounded corners=2.5pt, line width=0.35pt,
    text width=3.95cm, align=center, minimum height=7.2mm, inner ysep=0.8mm,
    drop shadow={shadow xshift=0.35pt, shadow yshift=-0.5pt, opacity=0.18, fill=black}},
  lab/.style={font=\scriptsize\bfseries, text=black!75, anchor=east},
  up/.style={-{Latex[length=1.1mm,width=1.1mm]}, black!55, line width=0.4pt},
  back/.style={{Latex[length=1.1mm,width=1.1mm]}-, black!45, line width=0.4pt, dashed, dash pattern=on 1.1pt off 1.1pt}
]
\node[band={black!40}{black!6}] (res) at (0,0)
  {\textbf{Typed results with provenance} \ $\cdot$ \ one comparison report};
\node[band={orange!55!black!45}{orange!12}, below=2.6mm of res] (con)
  {\textbf{Requirements declared and checked before execution}};
\coordinate (crow) at ($(con.south)+(0,-2.6mm)$);
\node[cell={blue!45!black!40}{blue!9}, anchor=north] (m) at ($(crow)+(-4.275cm,0)$)
  {Measure\\[0.2mm]\normalfont\footnotesize\bfseries 33 metrics};
\node[cell={green!40!black!40}{green!9}, anchor=north] (g) at (crow)
  {Mitigate\\[0.2mm]\normalfont\footnotesize\bfseries 14 mitigators};
\node[cell={violet!45!black!40}{violet!9}, anchor=north] (a) at ($(crow)+(4.275cm,0)$)
  {Audit\\[0.2mm]\normalfont\footnotesize\bfseries 14 diagnostics};
\node[band={black!40}{black!6}, below=2.6mm of g.south] (ev)
  {\textbf{Evidence containers} \ $\cdot$ \ \textbf{Model adapters}};
\foreach \n in {m,g,a}{\draw[up] (\n.north) -- (\n.north |- con.south);
                       \draw[up] (\n.south |- ev.north) -- (\n.south);}
\draw[up] (g.north |- con.north) -- (g.north |- res.south);
\draw[back] ($(ev.east)+(1.2mm,0)$) -- ($(ev.east)+(4.5mm,0)$) |- ($(res.east)+(1.2mm,0)$);
\node[lab] at ($(res.west)+(-1.5mm,0)$) {Results};
\node[lab] at ($(con.west)+(-1.5mm,0)$) {Contract};
\node[lab] at ($(m.west)+(-1.5mm,0)$) {Components};
\node[lab] at ($(ev.west)+(-1.5mm,0)$) {Evidence};
\end{tikzpicture}}
\vspace{-0.3cm}
\caption{Overview of the FairLMs design; the dashed arrow marks re-measurement.}
\label{fig:overview}
\vspace{-0.5cm}
\end{figure}

Figure~\ref{fig:overview} shows the four layers the components share. Evidence containers and model adapters supply the inputs; metrics, mitigators and diagnostics consume them under the declarations checked before execution; and every result carries its provenance into one comparison report. An intra-processing mitigator returns a model adapter, which re-enters the same path for re-measurement. The three component families differ in their declarations and in their return types, and are taken in turn below.

\sloppy
\noindent \textbf{Measurement.} Bias metrics require different evidence and model capabilities. WEAT~\citep{caliskan2017semantics} uses word embeddings; SEAT~\citep{may2019measuring} extends association testing to sentence representations, requiring templates and representation extraction; the original CrowS-Pairs score~\citep{nangia2020crows} uses masked-token probabilities; and the equal-opportunity gap~\citep{hardt2016equality} requires predictions, true labels, and group membership. Combining applicable metrics therefore requires coordinating evidence formats and model-specific tokenization and output conventions. Existing tools address parts of this problem: AIF360 and Fairlearn provide group-fairness metrics over task predictions~\citep{bellamy2019ai,weerts2023fairlearn}, LangFair supports output-based assessments~\citep{bouchard2025langfair}, and WEFE focuses on static word embeddings~\citep{badilla2025wefe}. Evaluation frameworks organize measurement through task or scenario configurations~\citep{eval-harness,liang2022holistic}, while \texttt{bias-bench} provides benchmark-specific pipelines~\citep{meade2022empirical}. FairPy~\citep{viswanath2023fairpy} consolidates several LM metrics but reports failures of some word-level probability metrics under subword tokenization. Evaluations spanning these interfaces still require reconciling model access, evidence formats, and metric-specific assumptions.

FairLMs addresses these requirements through explicit metric declarations and shared interfaces. Each registered metric declares supported architectures, required capabilities, and accepted evidence containers; model requirements are checked against the selected model profile before execution to identify mismatches with the declared requirements. Adapters encapsulate model-specific tokenization and output access for supported encoder-only, decoder-only, and encoder-decoder checkpoints and hosted completion APIs, so that compatible metrics reuse the same model integration. Sixteen built-in benchmark loaders~\citep[e.g.,][]{nangia2020crows,nadeem2021stereoset} and user-supplied data populate shared evidence containers, which reduces the format conversion between compatible metrics. For example, one \texttt{WordSets} object serves WEAT, SEAT, and a gradient-based attribution metric, while SEAT's templating and pooling remain metric-specific configuration choices. To make these choices inspectable, results record their configuration, together with the evidence hash and library version, so a user can trace each score to its evaluation settings.

\noindent \textbf{Bias Mitigation and Re-measurement.} Bias mitigation methods differ in intervention stage, required model access, and output type~\citep{zhang2026mitigating}. For example, counterfactual data augmentation~\citep{zhao2018gender} needs the training corpus and a lexicon of sensitive terms, but nothing from the model. Similarly, adversarial debiasing~\citep{elazar2018adversarial} needs control over the training objective and optimization, while group-aware thresholding~\citep{hardt2016equality} needs only scores, true labels, and group membership. Reusing them therefore means checking access requirements, integrating each output into the target system, and aligning evaluation configurations. Existing tools address parts of this problem: AIF360 and Fairlearn mitigate through dataset, estimator, and prediction interfaces~\citep{bellamy2019ai,weerts2023fairlearn}, WEFE debiases static embeddings~\citep{badilla2025wefe}, \texttt{bias-bench} provides pipelines for selected LM methods and checkpoints~\citep{meade2022empirical}, FairPy integrates several methods but reports difficulties accommodating differences in model representation layers~\citep{viswanath2023fairpy}, and the evaluation harnesses focus on assessing supplied models~\citep{eval-harness,liang2022holistic}. Workflows spanning these interfaces still require coordinating method requirements, returned objects, and evaluation configurations.

FairLMs connects requirement checking, intervention outputs, and re-measurement through a common contract. Each mitigator declares its intervention category, minimum access level, model requirements, and accepted evidence; compatibility checks assess the selected model and supplied evidence against the relevant requirements before execution. The return type is declared in the same way: a shared \texttt{apply} interface produces a \texttt{MitigationResult} holding transformed evidence or sample weights, a callable loss component, a \texttt{ModelAdapter}, or a fitted output rule, which determines the caller's remaining work. Evidence, weights, and loss components enter the caller's training procedure, output rules enter the prediction pipeline, and adapters are already usable by compatible metrics. Once that integration is complete, \texttt{compare\_before\_after} evaluates the baseline and the mitigated system under the same configured, compatible metrics, so that compatible interventions reuse a common evaluation workflow. Model-dependent quantities are computed through adapters, supplied as before/after evidence, or generated by a caller-provided evidence function; the caller must use the same evaluation instances on both sides.

\noindent \textbf{Dataset and Scoring-Instrument Diagnostics.} An observed bias score may reflect evaluation dataset design~\citep{blodgett2021stereotyping} and auxiliary scoring instruments~\citep{dixon2018measuring,jacobs2021measurement} as well as the target model. Documentation frameworks specify the information to be recorded about a dataset~\citep{gebru2021datasheets}, and published audits characterize individual benchmarks~\citep{blodgett2021stereotyping,zhang2025datasets}, but they report findings about the datasets they examine and do not provide tooling that a user can apply to another dataset. Existing toolkits offer related capabilities: AIF360 and Fairlearn support group-based analyses of labels and predictions~\citep{bellamy2019ai,weerts2023fairlearn}; FairPy, \texttt{bias-bench}, and WEFE evaluate model or embedding bias~\citep{viswanath2023fairpy,meade2022empirical,badilla2025wefe}, and LangFair assesses generated text using auxiliary scorers~\citep{bouchard2025langfair}. Diagnostics of the evaluation data and of the scoring instrument are not part of these interfaces, and assembling such an audit from them requires coordinating evidence interfaces and specifying dataset mappings, reference distributions, and diagnostic-specific decision rules.

FairLMs packages these audits as reusable components, grouped into representativeness, stereotype leakage, construction bias, and scoring-instrument behavior, that consume dataset evidence and scorer outputs without invoking the target LM, so a benchmark can be audited before any model is evaluated. To make the audit assumptions explicit, \texttt{DatasetAuditSpec} records the target, protected axes, and reference distributions, while diagnostic configurations specify extraction procedures, score thresholds, and auxiliary backends. Separating these choices from diagnostic implementations allows users to reuse the computations across datasets after mapping their evidence and adapting the audit settings as needed. Each diagnostic defines its estimand and declares the evidence and applicability conditions that are checked before it runs; component results distinguish \texttt{ready}, \texttt{blocked}, \texttt{not\_applicable}, and \texttt{failed}. The dataset-level estimators are designed to be comparable across
datasets~\citep{zhang2025datasets}, and the recorded specification and configuration then allow users to assess whether results share the measurement conditions needed for comparison. Scorer results are interpreted within the audited text and scoring context.

\vspace{-0.45cm}
\section{Quality Standards and Availability}
\label{sec:quality}
\vspace{-0.15cm}

\noindent \textbf{Correctness.} For the metric families with published reference values, we compare FairLMs against the publicly available reference implementation by running its configuration on the same checkpoint and evidence, and reproduce the published values within the precision at which they are reported. Golden fixtures pin every diagnostic to hand-derived values, and divergences from released implementations are documented with their reasons.

\noindent \textbf{Testing, documentation and openness.} A contract suite checks every registered component's declarations and applicability; the full suite runs in continuous integration, and its coverage report and a contribution guide are published with the MIT-licensed repository. Documentation covers installation, an API reference, eleven guides and a notebook.

\nocite{chu2024fairness,yin2026disentangled,wang2025history}

\vspace{-0.45cm}
\section{Conclusion}
\vspace{-0.15cm}

We have presented \texttt{FairLMs}, a Python library in which metrics, mitigators and diagnostics declare what they require of a model and of the evidence, so that compatible components can be combined, a mitigated system re-measured under the same configuration, and the datasets and scoring instruments behind a result audited. New components that follow the same declarations extend the library without changes to it.

\newpage

\vskip 0.2in
\bibliography{sample}

\end{document}